%% file: eurogp.tex
\documentclass{llncs}
\usepackage{subcaption}     %subfigures is outdated
\usepackage{graphicx}
\usepackage{latexsym}
\usepackage{amsxtra} 
\usepackage{amsmath}
\usepackage{amsfonts}
\usepackage{marvosym}
\usepackage{scalefnt}
\usepackage{color,tabularx}
\usepackage{makecell}
\usepackage{booktabs,multicol,multirow}
\usepackage{algorithm}
\usepackage{rotating}
\usepackage[noend]{algpseudocode}

  \usepackage[accepted]{llncsconf}

  \conference{24th European Conference on Genetic Programming (EuroGP)}

  \llncs{book editors and title}{0042} %% 0042 is the start page

\begin{document}
\title{Evolutionary Neural Architecture Search Supporting Approximate Multipliers \vspace*{-0.0em}}

\author{Michal Pinos, Vojtech Mrazek\orcidID{0000-0002-9399-9313}, and Lukas Sekanina\orcidID{0000-0002-2693-9011}}
\authorrunning{M. Pinos, V. Mrazek, and L. Sekanina}

\institute{Brno University of Technology, Faculty of Information Technology,\\IT4Innovations Centre of Excellence \\ 
Bo\v{z}et\v{e}chova 2, 612 66 Brno, Czech Republic\\
\email{ipinos@fit.vutbr.cz, mrazek@fit.vutbr.cz, sekanina@fit.vutbr.cz}} 

% make the title area
\maketitle

\begin{abstract}
There is a growing interest in automated neural architecture search (NAS) methods. They are employed to routinely deliver high-quality neural network architectures for various challenging data sets and reduce the designer’s effort. The NAS methods utilizing multi-objective evolutionary algorithms are especially useful when the objective is not only to minimize the network error but also to minimize the number of parameters (weights) or power consumption of the inference phase. We propose a multi-objective NAS method based on Cartesian genetic programming for evolving convolutional neural networks (CNN). The method allows approximate operations to be used in CNNs to reduce power consumption of a target hardware implementation. During the NAS process, a suitable CNN architecture is evolved together with approximate multipliers to deliver the best trade-offs between the accuracy, network size and power consumption. The most suitable approximate multipliers are automatically selected from a library of approximate multipliers. Evolved CNNs are compared with common human-created CNNs of a similar complexity on the CIFAR-10 benchmark problem. 

\end{abstract}

\section{Introduction}

Machine learning technology based on deep neural networks (DNNs) is currently penetrating into many new application domains. It is deployed in such classification, prediction, control, and other tasks in which designers can collect comprehensive data sets that are mandatory for training and validating the resulting model. In many cases (such as smart glasses or voice assistants), DNNs have to be implemented in low-power hardware operated on batteries. Particularly, the inference process of a fully trained DNN is typically accelerated in hardware to meet real-time time requirements and other constraints. Hence, drastic optimizations and approximations have to be introduced at the level of hardware~\cite{Capra:cnn:hw:survey:2020}. On the other hand, DNN training is typically conducted on GPU servers.

Existing DNN architectures have mostly been developed by human experts \emph{manually}, which is a time-consuming and error-prone process. The current approach to hardware implementations of DNNs is based on semi-automated simplifying of a network model, which was initially developed for a GPU and trained on GPU without considering any hardware implementation aspects. There is a growing interest in automated DNN design methods known as the \emph{neural architecture search} (NAS)~\cite{JMLR:v20:18-598,Wistuba:19}.  Evolutionary NAS, introduced over three decades ago~\cite{yao:pieee99}, is now intensively adopted, mostly because it can easily be implemented as a multi-objective design method~\cite{Stanley:nature:2019,Lu:nsganet:gecco19}.

This paper is focused on the NAS applied to the automated design of \emph{convolutional neural networks} (CNNs) for image classification. Current NAS methods only partly reflect hardware-oriented requirements on resulting CNNs. In addition to the classification accuracy, some of them try to minimize the number of parameters (such as multiply and accumulate operations) for a GPU implementation, which performs all operations in the floating-point number representation~\cite{Lu:nsganet:gecco19,JMLR:v20:18-598}.
Our research aims to propose and evaluate a NAS method for the highly automated design of CNNs that reflect hardware-oriented requirements. We hypothesize that \emph{more energy-efficient hardware implementations of CNNs can be obtained if hardware-related requirements are specified, reflected, and exploited during the NAS}. In this paper, we specifically focus on the automated co-design of CNN’s topology and approximate arithmetic operations. The objective is to automatically generate CNNs showing good trade-offs between the accuracy, the network size (the number of multiplications), and a degree of approximation in the used multipliers. 
%Other hardware-related aspects are left for future studies.

The proposed method is based on a multi-objective Cartesian genetic programming (CGP) whose task is to maximize the classification accuracy and minimize the power consumption of the most dominated arithmetic operation, i.e., multiplications conducted in convolutional layers. To avoid the time-consuming automated design of approximate multipliers, CGP selects suitable multipliers from a library of approximate multipliers~\cite{mrazek:date:17}. While CGP delivers the network topology, the weights are obtained using a TensorFlow. 
%standard learning method. 
The NAS supporting approximate multipliers in CNNs is obviously more computationally expensive than the NAS of common CNNs. The reason is that TensorFlow does not support the fast execution of CNNs that contain non-standard operations such as approximate multipliers. 
We propose eliminating this issue by employing TFApprox~\cite{Vaverka:date:2020}, which extends TensorFlow to support approximate multipliers in CNN training and inference. Evolved CNNs are compared with common human-created CNNs of a similar complexity on the CIFAR-10 benchmark problem.

To summarize our key contributions:
We present a method capable of an automated design of CNN topology with automated selection of suitable approximate multiplier(s). The methodology uniquely integrates a multi-objective CGP and TFApprox-based training and evaluation of CNNs containing approximate circuits.
We demonstrate that the proposed method provides better trade-offs than a common approach based on introducing approximate multipliers to CNNs developed without reflecting any hardware aspects.

%The rest of the paper is organized as follows. Section~\ref{sec:soa} briefly surveys relevant work dealing with NAS, CGP, and approximate multipliers. Section~\ref{sec:method} introduces the proposed CGP-based method for CNNs utilizing approximate multipliers. The experimental setup, results, and comparisons with other relevant works are presented in Section~\ref{sec:results}. Conclusions are given in Section~\ref{sec:conclusions}.

\section{Related Work}
\label{sec:soa}

Convolutional neural networks are deep neural networks employing, in addition to other layer types, the so-called \emph{convolutional layers}. 
These layers are capable of processing large input vectors (tensors). Simultaneously, the number of parameters (the weights in the convolutional kernels) they use is small compared to the common fully-connected layers. 
Because the state of the art CNNs consist of hundreds of layers and millions of network elements, they are demanding in terms of the execution time and energy requirements. For example, the inference phase of a trained CNN such as ResNet-50 requires performing $3.9 \cdot 10^9$ multiply-and-accumulate operations to classify one single input image. Depending on a particular CNN and a hardware platform used to implement it, arithmetic operations conducted in the inference are responsible for 10\% to 40\% of total energy~\cite{sze:pieee17}. 

To reduce power consumption, hardware-oriented optimization techniques developed for CNNs focus on optimizing the data representation, pruning less important connections and neurons, approximating arithmetic operations, compression of weights, and employing various smart data transfer and memory storage strategies~\cite{jiang2020standing,Panda:dnn16}. For example, the Ristretto tool is specialized in determining the number of bits needed for arithmetic operations~\cite{Ristretto} because the standard 32-bit floating-point arithmetic is too expensive and unnecessarily accurate for CNNs. According to~\cite{sze:pieee17}, an 8-bit fixed-point multiply consumes 15.5$\times$ less energy (12.4$\times$ less area) than a 32-bit fixed-point multiply,
and 18.5$\times$ less energy (27.5$\times$ less area) than a 32-bit floating-point multiply. Further savings in energy are obtained not only by the bit width reduction of arithmetic operations but also by introducing approximate operations, particularly to the multiplication circuits~\cite{Mrazek:iccad:2019,Sarwar:2018}. Many approximate multipliers are available in public circuit libraries, for example, EvoApproxLib~\cite{mrazek:date:17}. All these techniques are usually applied to CNN architectures initially developed with no or minimal focus on a potential hardware implementation.

NAS has been introduced to automate the neural network design process. The best-performing CNNs obtained by NAS currently show superior performance with respect to human-designed CNNs~\cite{JMLR:v20:18-598,Wistuba:19}. NAS methods can be classified according to the \emph{search mechanism} that can be based on reinforcement learning~\cite{DBLP:journals/corr/ZophL16}, evolutionary algorithms (EA)~\cite{neat}, gradient optimization~\cite{DBLP:journals/corr/abs-1806-09055}, random search~\cite{RandomSearch}, or sequential model-based optimization~\cite{PNAS}. NAS methods were initially constructed as single-objective methods to minimize the classification error for a CNN running on a GPU~\cite{largescale,Suganuma:GECCO2017}. Recent works have been devoted to \emph{multi-objective} NAS approaches in which the error is optimized together with the cost, whose minimizing is crucial for the sustainable operation of GPU clusters~\cite{MONAS,Lu:nsganet:gecco19}.

As our NAS method employs genetic programming, which is a branch of evolutionary algorithms, we briefly discuss the main components of the EA-based approaches. Regarding the \emph{problem representation}, direct~\cite{Lu:nsganet:gecco19,Suganuma:GECCO2017} and indirect (generative)~\cite{Stanley:nature:2019} encoding schemes have been investigated. The selection of \emph{genetic operators} is tightly coupled with the chosen problem representation. While mutation is the key operator for CGP~\cite{Suganuma:GECCO2017}, the crossover is crucial for binary encoding of CNNs as it allows population members to share common building-blocks~\cite{Lu:nsganet:gecco19}.
%Some methods use advanced techniques such as Bayesian learning~\cite{}.
The \emph{non-dominated sorting}, known from, e.g., the NSGA-II algorithm~\cite{deb2002}, enables to maintain diverse trade-offs between conflicting design objectives. The evolutionary search is often combined with \emph{learning} because it is very inefficient to let the evolution find the weights. A candidate CNN, constructed using the information available in its genotype, is trained using common learning algorithms available in popular DNN frameworks such as TensorFlow. The number of epochs and the training data size are usually limited to reduce the training time, despite the fact that by doing so the fitness score can wrongly be estimated. The CNN accuracy, which is obtained using test data, is interpreted as the fitness score. 
%If the second objective is to minimize the network cost, the most commonly used metrics are the number of parameters (weights), FLOPS, or multiplications. 
The best-evolved CNNs are usually \emph{re-trained} (fine-tuned) to further increase their accuracy.

The entire \emph{neuro-evolution} is very time and resources demanding and, hence, only several hundreds of candidate CNNs can be generated and evaluated in one EA run. On common platforms, such as TensorFlow, all mathematical operations are highly optimized and work with standard floating-point numbers on GPUs. If one needs to replace these operations with approximate operations, these non-standard operations have to be expensively emulated. The CNN execution is then significantly slower than with the floating-point operations. This problem can partly be eliminated by using TFApprox in which all approximate operations are implemented as look-up tables and accessed through a texture memory mechanism of CUDA capable GPUs~\cite{Vaverka:date:2020}. 

%With TFApprox, the inference time can significantly be reduced in these cases and in comparison with common TensorFlow~\cite{Vaverka:date:2020}.

A very recent work~\cite{jiang2020standing} presents a method capable of jointly searching the neural architecture, hardware architecture, and compression model for FPGA-based CNN implementations. 
%The search, based on reinforcement learning, starts from a preselected set of trained models to reduce the search time, and uses several optimization techniques including pattern pruning, channel pruning, filter expansion, and quantization. 
Contrasted to our work, arithmetic operations are performed on 16 bits, and no approximate operations are employed. High-quality results are presented for CIFAR-10 and ImageNet benchmark data sets.

\section{Evolutionary NAS with Approximate Circuits}
\label{sec:method}

The proposed evolutionary NAS is inspired in paper~\cite{Suganuma:GECCO2017} whose authors used CGP to evolve CNNs. We extend this work by (i) supporting a multi-objective search, (ii) using an efficient seeding strategy and (iii) enabling the approximate multipliers in convolutional layers. The method is evaluated on the design of CNNs for a common benchmark problem -- the CIFAR-10 image classification data set~\cite{CIFAR-10}. The role of CGP is to provide a good CNN architecture. The weights are obtained using Adam optimization algorithm implemented in TensorFlow. Our ultimate goal is to deliver new CNN architectures that are optimized for hardware accelerators of CNNs in terms of the parameter count and usage of low-energy arithmetic operations. 
%We assume that such an accelerator employs a two-dimensional array of processing elements (PE). A typical PE multiplies the input with its weight and updates the sum maintained in each layer. The PE array can be operated in several ways, see~\cite{sze:pieee17}; we suppose that the CNN accelerator is organized as a generic PE array and the \emph{same} multiplier is always used in all convolutional layers.

In this section, we will describe the proposed CGP-based NAS which is developed for CNNs with floating-point arithmetic operations executed on GPU. In Section~\ref{sec:approx:mult:cnn}, the proposed evolutionary selection of approximate multipliers for CNNs will be presented.   

\subsection{CNN Representation}

CGP was developed to automatically design programs and circuits that are modeled using directed acyclic graphs~\cite{miller:cgp:book}. A candidate solution is represented using a two-dimensional array of $n_c \times n_r$ nodes, consuming $n_i$ inputs and producing $n_o$ outputs. In the case of evolutionary design of CNNs, each node represents either one layer (e.g., fully connected, convolutional, max pooling, average pooling) or a module (e.g., residual or inception block) of a CNN. Each node of $j$-th column reads a tensor coming from column $1, 2, \dots, j-1$ and produces another tensor. In our case study, CNNs accept one 4D input tensor (of shape [batch\_size, height, width, depth]) holding one batch of input images and produce one 2D output tensor (of shape [batch\_size, class\_probs]) which is treated as a matrix, in which each row corresponds to a vector of class probabilities. 

\begin{figure}
    \centering
  \includegraphics[width=0.7\textwidth]{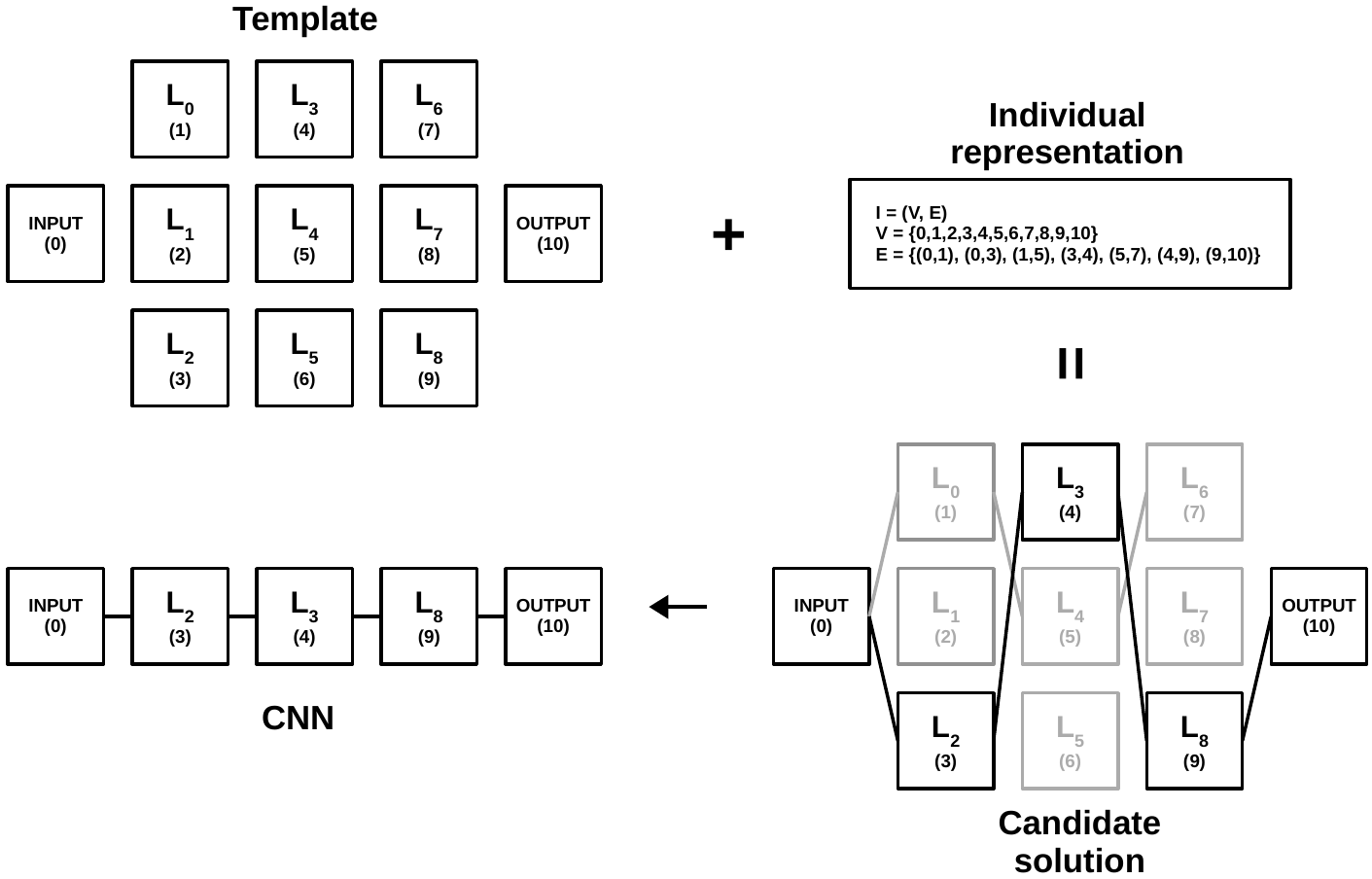}
  \caption{A template combined with an individual representation creates a candidate solution, which is transformed into a CNN.}
  \label{fig:cgp2cnn}
\end{figure}

Fig.~\ref{fig:cgp2cnn} shows how resulting CNN is obtained from an array of CGP nodes (called the  template) and an individual, which is represented as a graph $I = (V, E)$, where $V$ denotes a set of vertices (nodes of template) and $E$ is a set of edges. Individual representation (in a form of the graph $I$) in a conjunction with the template creates a candidate solution. The nodes that are employed in the CNN are called the active nodes and form a directed acyclic graph (DAG), that connects the input node with the output node. When a particular CNN has to be built and evaluated, this DAG is extracted (from a candidate solution) and transformed to a computational graph which is processed by TensorFlow. 

\begin{figure}
     \centering
     \begin{subfigure}[b]{0.3\textwidth}
         \centering
         \includegraphics[width=\textwidth]{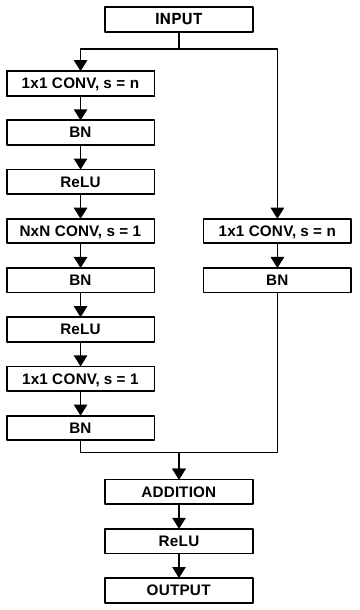}
         \caption{Bottleneck.}
         \label{fig:residual_bottleneck}
     \end{subfigure}
     \hfill
     \begin{subfigure}[b]{0.3\textwidth}
         \centering
         \includegraphics[width=\textwidth]{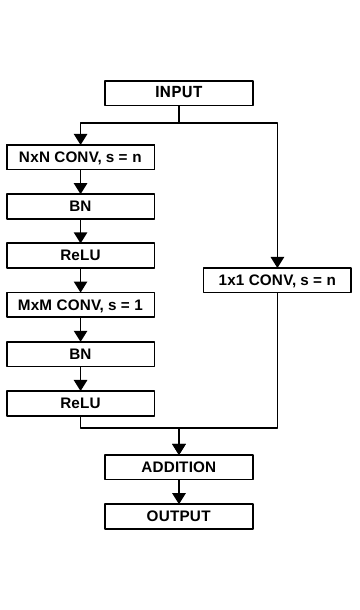}
         \caption{Residual.}
         \label{fig:residual}
     \end{subfigure}
     \hfill
     \begin{subfigure}[b]{0.3\textwidth}
         \centering
         \includegraphics[width=\textwidth]{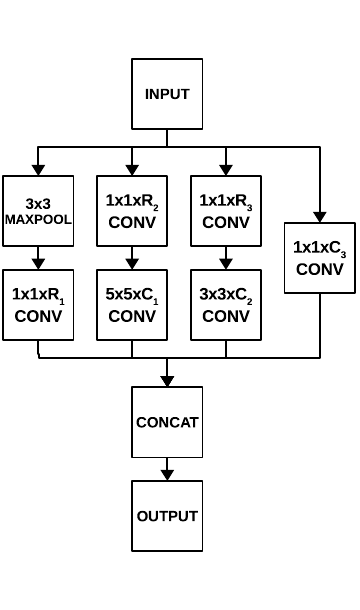}
         \caption{Inception.}
         \label{fig:inception}
     \end{subfigure}
     
        \caption{Diagrams of (a) bottleneck residual module, (b) residual module and (c) inception module.}
        \label{fig:modules}
\end{figure}

The following layers are supported: fully connected (FC), convolutional (CO-NV), summation (SUM), maximum pooling (MAX) and average pooling (AVG). Inspired in~\cite{Xie:iccv17}, CGP can also use inception (INC), residual (RES), and (residual) bottleneck (RES-B) modules~\cite{Kaiming:2016} that are  composed of several elementary layers as shown in Fig.~\ref{fig:modules}. Selected layers and modules are introduced in the following paragraphs; the remaining ones are standard.

%%%%%%%% Michaluv puvodni popis  %%%%%%%

The \emph{summation layer} accepts tensors $t_{1}$ and $t_{2}$ with $shape(t_{1}) = (h_{1}, w_{1}, c_{1})$ and $shape(t_{2}) = (h_{2}, w_{2}, c_{2})$, where $h_{x}, w_{x}$ and $c_{x}$ are height, width and the number of channels respectively ($x \in \{1, 2\}$). It outputs $t_{o}$, i.e. the sum of $t_{1}$ and $t_{2}$, defined as
\begin{align}
t_{o} = t_{1} + t_{2} \iff t_{o}^{ijk} = t_{1}^{ijk} + t_{2}^{ijk}\text{ for  }i &= 0, ..., c-1\\j &= 0, ..., h-1\nonumber\\k &= 0, ..., w-1 \nonumber. 
\end{align}
It has to be ensured that the height and width of both the tensors are identical. If it is not so, the pooling algorithm is applied to the `bigger’ tensor to unify these dimensions. The problem with unmatched number of channels is resolved by zero padding applied to the `smaller’ tensor, i.e., $shape(t_{o}) = (h_{o}, w_{o}, c_{o})$, where $h_o = min(h_{1}, h_{2})$, $w_{o} = min(w_{1}, w_{2})$ and $c_{o} = max(c_{1}, c_{2})$.

The \emph{inception module}, showed in Fig.~\ref{fig:inception}, performs in parallel three convolutions with filters 5x5, 3x3 and 1x1 and one maximum pooling. The results are then concatenated along the channel dimension. Additionally, 1x1 convolutions are used to reduce the number of input channels. Parameters $C_1$, $C_2$ and $C_3$ correspond to the number of filters in 5x5, 3x3 and 1x1 convolutions, whereas $R_1$, $R_2$ and $R_3$ denote the number of filters in 1x1 convolutions. All convolutional layers operate with stride 1 and are followed by the ReLU activation.

The \emph{residual module} contains a sequence of NxN and MxM convolutions that can be skipped, which is implemented by the summation layer followed by the ReLU activation. The residual module, shown in Fig.~\ref{fig:residual}, consists of two convolutional layers with the filters NxN and MxM, both followed by batch normalization and ReLU activation. In parallel, one convolution with filter 1x1 is computed. Results of MxM and 1x1 convolution are added together to form a result. Convolutional layers with filters NxN and 1x1 operate with stride $n$.

We also support a bottleneck variant of the residual module, shown in Fig.~\ref{fig:residual_bottleneck}, which comprises of one convolutional layer with filter NxN, which applies batch normalization and ReLU activation to its input and output. This convolutional layer is surrounded by two 1x1 convolutional layers. In parallel, another 1x1 convolutional layer is employed. The first two parallel 1x1 convolutional layers operate with stride $n$, whereas all other layers use stride $1$. The outputs of the last two parallel 1x1 convolutional layers are then batch-normalized and added together. The final output is obtained by application of ReLU activation to the output of the addition layer.

%%%%%%%%%%%%%%%%%

\subsection{Genetic Operators}

CGP usually employs only one genetic operator -- mutation. The proposed mutation operator modifies architecture of a candidate CNN; however, the functionality (layers and their parameters) implemented by the nodes are not directly changed, except some specific cases, see below. A randomly selected node is mutated in such a way that all its incoming edges are removed and a new connection is established to a randomly selected node situated in up to $L$ previous columns, where $L$ is a user-defined parameter. This is repeated $k$ times, where $k$ is the node’s arity. If the mutation does not hit an active node it is repeated to avoid generating functionally identical networks. One mutation can thus modify several inactive nodes before finally modifying an active node. 
%This is an important property of CGP\michal{, called \emph{neutral genetic drift} which provides a mechanism to escape from local optima~\cite{miller:cgp:book}}. 
The weights associated with a newly added active node are randomly initialized. If the primary output undergoes a mutation, its destination is a randomly selected node of the last column containing FC layers.

%\ls{In order to minimize the time-consuming training, the weights associated with the active nodes are directly inherited by their offspring. A potential problem is that the offspring’s node with the inherited weights is connected to a different node than in its parent. If dimensions of tensors specifying two nodes that have to be connected do not match each other, one of them is either cut or padded with randomly generated values to establish a correct connection in the offspring.}

\subsection{Fitness Functions}

The objectives are to maximize the CNN accuracy and to minimize the CNN complexity (which is expressed as the number of parameters), and power consumption of multiplication in the convolutional layers.
The objective function expressing the accuracy of a candidate network $x$ (evaluated using a data set $D$), is calculated using TensorFlow as
%\begin{equation}
%    \label{f1}
$f_1(x, D) = accuracy (x, D).$
%\end{equation}
The number of parameters in the entire CNN $x$ is captured by fitness function $f_2(x)$.
%\begin{equation}
%    \label{f1}
%f_2(x) = parameters\_count (x).
%\end{equation}
Power consumption is estimated as
%\begin{equation}
%    \label{f3}
$f_3(x) = N_{mult}(x) \cdot P_{mult},$
%\end{equation}
where $N_{mult}$ is the number of multiplications executed during inference in all convolutional layers and $P_{mult}$ is power consumption of used multiplier.

\subsection{Search algorithm}

\begin{figure}
    \centering
  \includegraphics[width=0.8\textwidth]{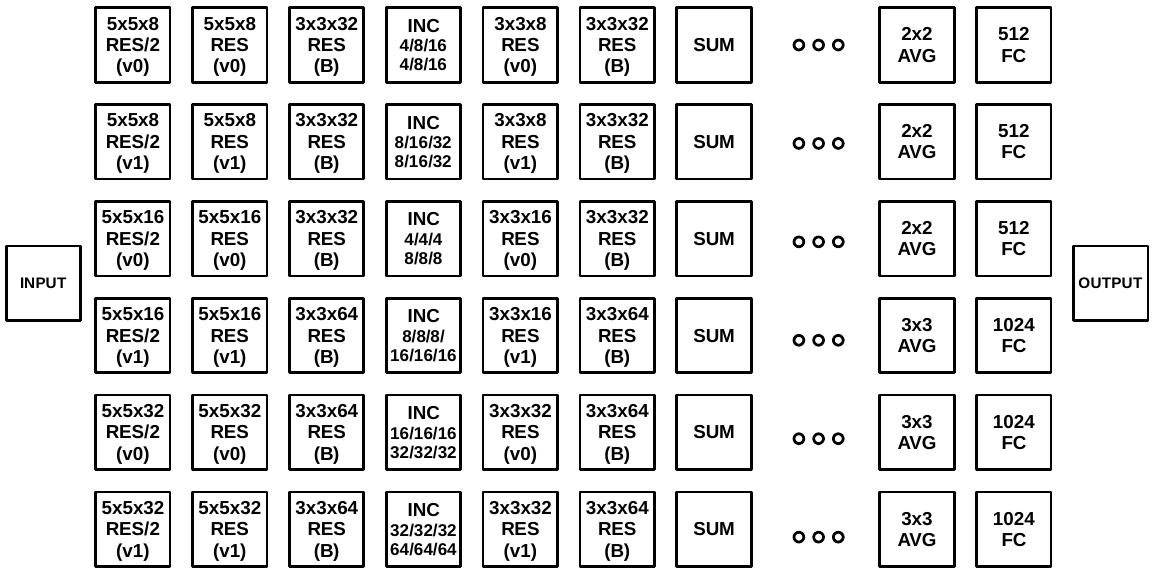}
  \caption{The template used to initialize CGP. 
%  Each node is described by parameters for specified layers or modules they represent. Residual modules RES (v0) and RES (B) and inception modules INC ($R_n;C_n$) correspond to the modules, shown in figure \ref{fig:modules}. Residual module (v1) is essentially same as (v0), except it applies batch normalization and ReLU activation before it performs the convolutions.
}
  \label{fig:template}
\end{figure}

The search algorithm (see Alg. 1) is constructed as a multi-objective evolutionary algorithm inspired in CGP-based NAS~\cite{Suganuma:GECCO2017} and NSGA-II~\cite{deb2002}. The initial population is heuristically initialized with networks created according to a template shown in Fig.~\ref{fig:template}. 
The template consists of typical layers of CNNs, i.e., convolutional layers in the first and middle parts and fully connected layers at the end. 
All connections in the template (including the link to the output tensor) and all associated weights are randomly generated. The proposed template ensures that even the networks of the initial populations are reasonable CNNs which reduces the computational requirements of the search process.

Training of a CNN is always followed by testing to obtain fitness values $f_1(x)$, $f_2(x)$, and $f_3(x)$. To reduce the training time, a randomly selected subset $D_{train}$ of the training data set can be used. The same subset is used for training all the individuals belonging to the same population. Training is conducted for $E_{train}$ epochs. The accuracy of the candidate CNN (i.e., $f_1$) is determined using the entire test data set $D_{test}$ (Alg. 1, line 2). To overcome the overfitting during the training, data augmentation and L2 regularization  were employed \cite{Augmentation}.

The offspring population ($O$) is created by applying the mutation operator on each individual of the parental population $P$. The offspring population is evaluated in the same way as the parental population (Alg. 1, line 5). Populations $P$ and $O$ are joined to form an auxiliary population $R$ (line 6). The new population is constructed by selecting non-dominated individuals from Pareto fronts (PF) established in $R$ (lines 9 -- 10). If any front must be split, a crowding distance is used for the selection of individuals to $P$ (lines 12 -- 13)~\cite{deb2002}. The search terminates after evaluating a given number of CNNs.

As the proposed algorithm is multi-objective, the result of a single CGP run is a set of non-dominated solutions. At the end of evolution, the best-performing individuals from this set are re-trained (fine-tuned) for $E_{retrain}$ epochs on the complete training data set $D_{retrain}$ and the final accuracy is reported on the complete test data set $D_{test}$.

\begin{algorithm}[h!]
	\caption{Neuroevolution}
    \input{nas_alg}
	\label{alg:NAS}
\end{algorithm}

\subsection{NAS with Approximate Multipliers}
\label{sec:approx:mult:cnn}

So far we have discussed a NAS utilizing standard floating-point arithmetic operations. In order to find the most suitable approximate multiplier for a CNN architecture, we introduce the following changes to the algorithm. (i) The problem representation is extended with one integer specifying the index $I_m$ to the list of available 8-bit approximate multipliers, i.e. to one of the 14 approximate multipliers included in EvoApproxLib-Lite\footnote{http://www.fit.vutbr.cz/research/groups/ehw/approxlib/}~\cite{mrazek:date:17}. These approximate multipliers show different trade-offs between power consumption, error metrics and other parameters. Please note that the selection of the exact 8-bit multiplier is not excluded. (ii) The mutation operator is modified to randomly change this index with probability $p_{mult}$. (iii) Before a candidate CNN is sent to TensorFlow for training or testing, all standard multipliers used in convolutional layers are replaced with the 8-bit approximate multiplier specified by index $I_m$. TensorFlow, with the help of TFApprox, then performs all multiplications in the convolution computations in the forward pass of learning algorithm with the approximate multipliers, whereas all computations in the backward pass are done with the standard floating-point multiplications.

\section{Results}
\label{sec:results}

Table~\ref{tab:params} summarizes all parameters of CGP and the learning method used in our experiments. These parameters were experimentally selected based on a few trial runs. Because of limited computational resources, we could generate and evaluate only ${pop}\_{size} + G \times {pop}\_{size} = 88$ candidate CNNs in each run. 

\begin{table}[ht]
    \centering
    \tabcolsep 10pt
    \caption{Parameters of the experiment.}
    \label{tab:params}
    \begin{tabular}{l c l}\toprule
         \bf Parameter & \bf Value  & \bf Description\\
         \midrule
         $n_r$ & 6  & Number of rows in the CGP grid.\\
         $n_c$ & 23 & Number of columns in the CGP grid.\\
         $L$ & 5 & L-back parameter.\\
         ${pop}\_{size}$ & 8 & Number of individuals in the population.\\
         $G$ & 10 & Maximum number of generations.\\
         \multirow[t]{2}{*}{$D_{train}$} & 50000 & Size of the data set used during the evolution. \\ 
         %& & the evaluation of the individuals.\\
         \multirow[t]{2}{*}{$D_{retrain}$} & 50000 & Size of the data set for re-training. \\
         %& & the final retraining of the best individuals.\\
         $D_{test}$ & 10000 & Size of the test data set.\\
         \multirow[t]{2}{*}{$E_{train}$} & 20 & Number of epochs (during the evolution).\\
         %& &  during the evaluation of the individuals.\\
         \multirow[t]{2}{*}{$E_{retrain}$} & 200 & Number of epochs (for re-training).\\
        % & &  during the final training of the best individuals.\\
         \multirow[t]{2}{*}{$batch\_size$} & 32 & Batch size.\\ %Number of %samples, used to feed the CNN\\
         %& & model in one iteration of the training.\\
         $rate$ & 0.001 & Initial learning rate for all CNNs.\\
         \multirow[t]{2}{*}{$p_{arch}$} & 1.0 & Probability of mutation of the architecture.\\
         %& & individual will be mutated.\\
         \multirow[t]{2}{*}{$p_{mult}$} & 1.0 & Probability of mutation of $I_m$ \\
         % & &  individual will be mutated.\\
    
         \bottomrule
    \end{tabular}
\end{table}

\subsection{The Role of Approximate Multipliers in NAS}

We consider four scenarios to analyze the role of approximate multipliers in NAS: (S1) CNN is co-optimized with the approximate multiplier under fitness functions $f_1$ and $f_3$ (denoted `CGP+auto-selected-mult-A/E' in the following figures); (S2) CNN is co-optimized with the approximate multiplier under fitness functions $f_1, f_2$ and $f_3$ (denoted `CGP+auto-selected-mult-A/E/P'); (S3) A selected approximate multiplier is always used in NAS (denoted `CGP+fixed-approx-mult-A/E'); (S4) The 8-bit exact multiplier is always used in NAS (denoted `CGP+accurate-8-bit-mult-A/E'). Note that symbols A, E and P denote Accuracy, Parameters and Energy.
Because of limited resources we executed 5, 2, 13, and 2 CGP runs for scenarios S1, S2, S3, and S4.

\begin{figure}
    \centering
    \includegraphics[width=\columnwidth]{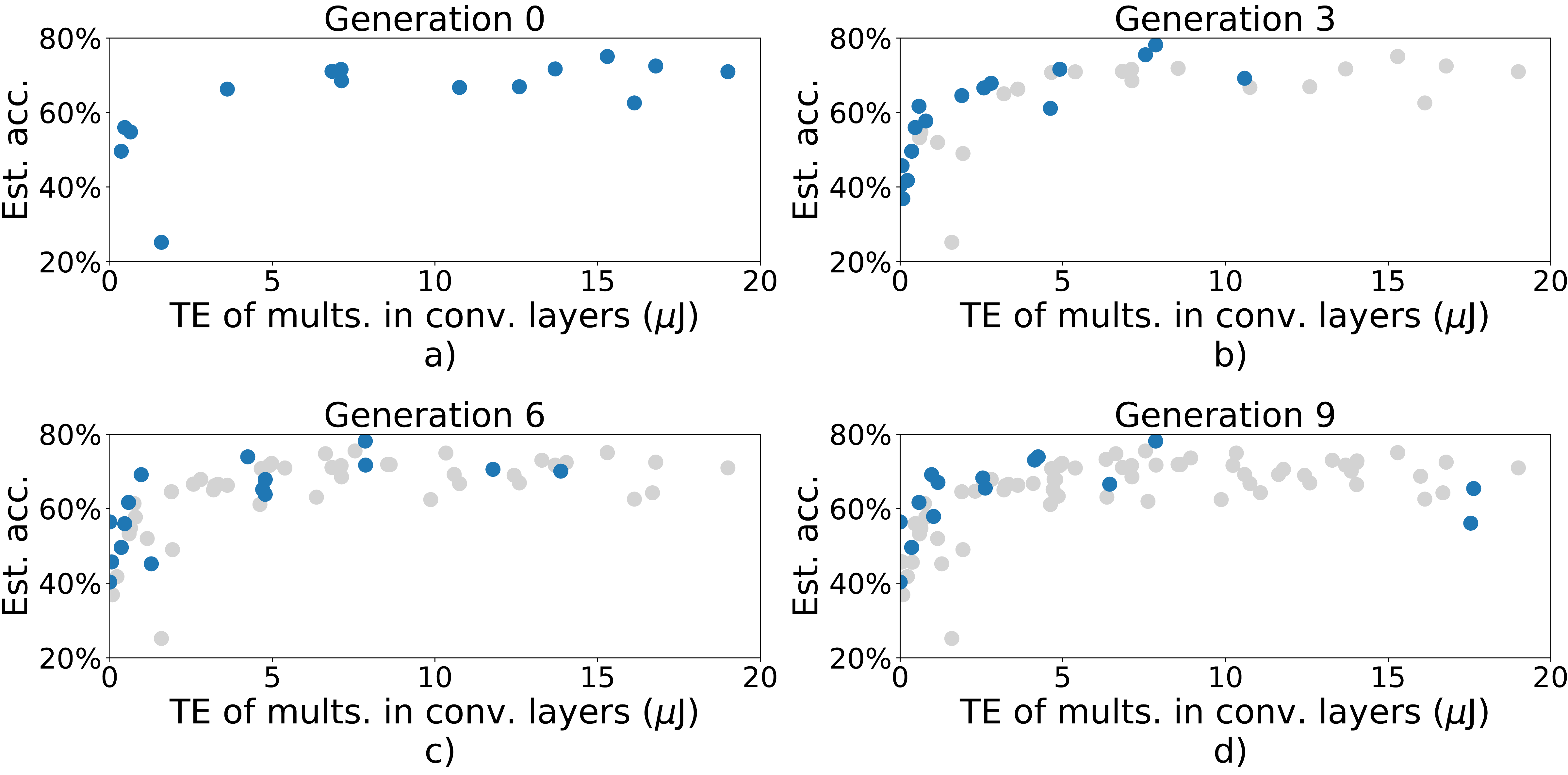}
    \caption{A typical progress of evolution in scenario S1 (Estimated Accuracy vs. Total Energy). The blue points represent the current generation (all parents and offspring). The grey points are all previously generated solutions.}
    \label{fig:gen}
\end{figure}

Fig.~\ref{fig:gen} plots a typical progress of a CGP run in S1 scenario. The blue points represent the initial population -- all parents and offspring are depicted in Fig.~\ref{fig:gen}a. The remaining subfigures show generations 3, 6, and 9. The grey points are candidate solutions created in the previous generations and their purpose is to emphasize the CGP progress. As the best trade-offs (Accuracy vs. Energy) are moving to the top-left corner of the figures, we observe that CGP can improve candidate solutions despite only 10 populations are generated.

\begin{figure}
    \centering
    \includegraphics[width=\columnwidth]{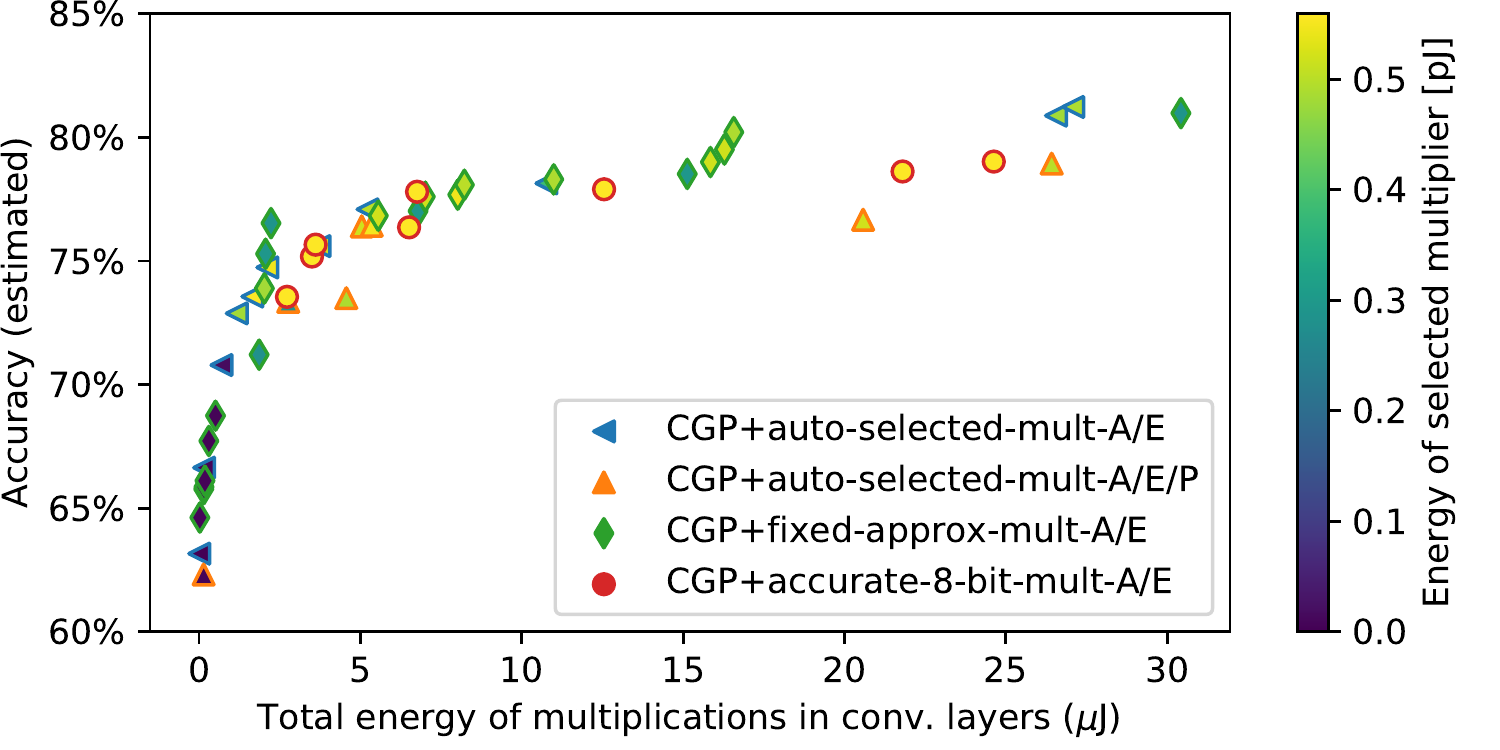}
    \caption{Trade-offs between the accuracy and the total energy of multiplications performed in convolutional layers during one inference obtained with different design scenarios.}
    \label{fig:multcost}
\end{figure}

\begin{figure}
    \centering
    \includegraphics[width=\columnwidth]{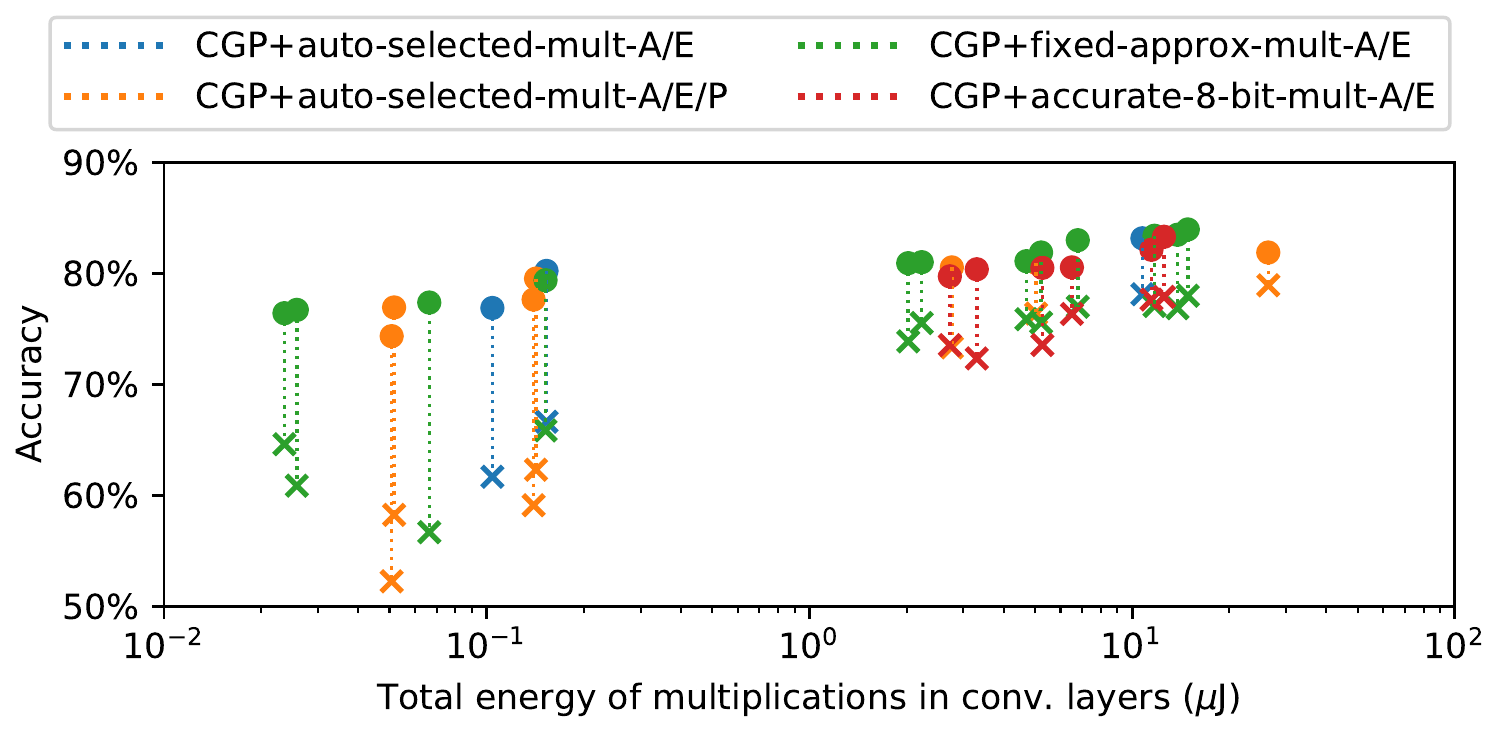}
    \caption{The impact of re-training on the accuracy of best-evolved CNNs. Crosses/points denote the accuracy before/after re-training.}
    \label{fig:retraining}
\end{figure}

Trade-offs between the accuracy (estimated in the fitness function) and the total energy of multiplications performed in convolutional layers during inference are shown in Fig.~\ref{fig:multcost}. The Pareto front is mostly occupied by CNNs evolved in scenario S3, i.e. with a pre-selected approximate multiplier. CNNs utilizing the 8-bit accurate multiplier are almost always dominated by CNNs containing some of the approximate multipliers. CNNs showing the 70\% and higher accuracy never use highly approximate multipliers (see the bar on the right hand side) in Fig.~\ref{fig:multcost}. Fig.~\ref{fig:retraining} shows that re-training of the best evolved CNNs conducted for $E_{retrain}$ epochs significantly improves the accuracy.

\begin{figure}
    \centering
    \includegraphics[width=0.9\columnwidth]{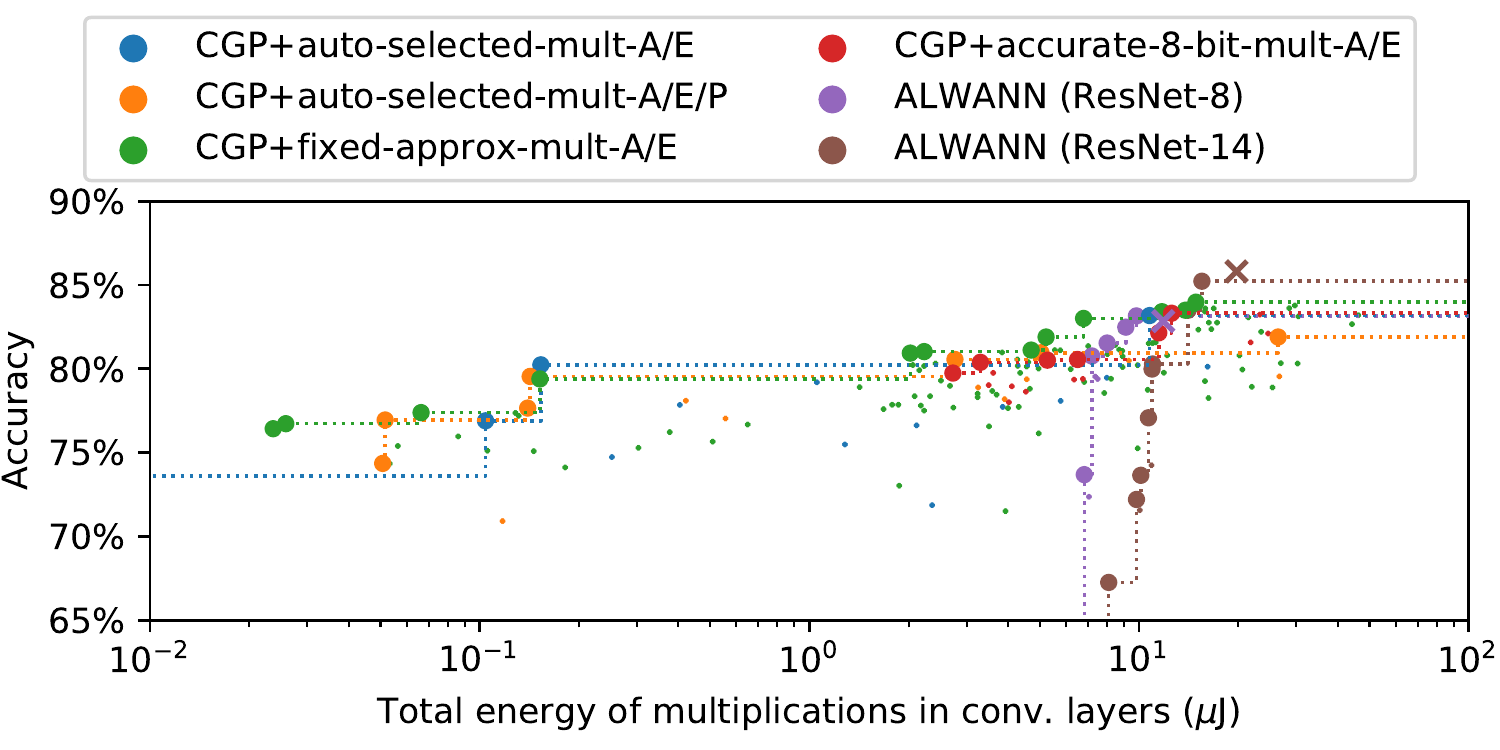}
    \caption{Pareto fronts obtained in four scenarios compared with ResNet networks utilizing 8-bit multipliers (crosses) and the ALWANN method.}
    \label{fig:final}
\end{figure}

Final Pareto fronts obtained (after re-training) in our four scenarios are highlighted in Fig.~\ref{fig:final}. When an approximate multiplier is fixed before the NAS is executed (S3), CGP is almost always able to deliver a better trade-off than if a suitable multiplier is automatically selected during the evolution (in S1 or S2). However, CGP has to be repeated with each of the pre-selected multipliers to find the best trade-offs. We  hypothesize that longer CGP runs are needed to benefit from S1 and S2. 

\begin{figure}
    \centering
    \includegraphics[width=0.6\columnwidth]{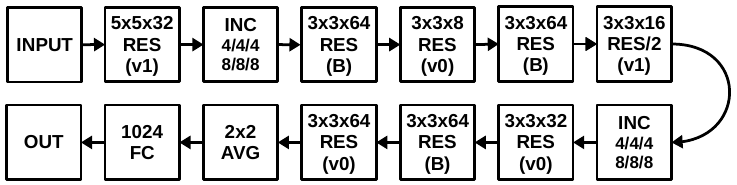}
    \caption{Evolved CNN whose parameters are given in the first row of   Table~\ref{tab:summary}.}
    \label{fig:selected:CNN}
\end{figure}

Finally, Table~\ref{tab:summary} lists key parameters of selected CNNs (the final and estimated accuracy, the energy needed for all multiplications in convolutional layers, and the number of multiplications) and used approximate multipliers (the identifier and the energy per one multiplication). One of the evolved CNNs is depicted in Fig.~\ref{fig:selected:CNN}. 
%\ls{This CNN employs mul8\_QJD approximate multiplier (see EvoApproxLib) which shows a 12\% reduction in power consumption compared with an exact multiplier. Its Error Probability is 75\% and the Worst Case Error is 0.082\%.}

%\ls{The data created during our experiments were also used to analyse a correlation between the CNN accuracy and the error of the approximate multiplier used by CNN. We revealed that the CNN accuracy highly correlates with the Mean Relative Error of the multiplier (Pearson coefficient = 0.814). Correlations with the Mean Absolute Error (Pearson coefficient = 0.717) and the Error Probability (Pearson coefficient = 0.550) are less significant.}

\begin{table}[ht]
    \centering
    \caption{Key parameters of selected CNNs and used multipliers. Symbol ${}^{*}$ denotes the 8-bit accurate multiplier.}
    %\resizebox{\columnwidth}{!}{\input{tab_all}}
    {\input{tab_all}
}
    \label{tab:summary}
\end{table}

\subsection{Execution Time}
Experiments were performed on a machine with two 12-core CPUs Intel Skylake Gold 6126, 2.6 GHz, 192 GB, equipped with four GPU accelerators NVIDIA Tesla V100-SXM2. A single CGP run with CNNs utilizing approximate multipliers takes 48 GPU hours; the final re-training requires additional 56 GPU hours on average. 
When approximate multipliers are emulated by TFApprox, the average time needed for all inferences in ResNet-8 on CIFAR-10 is 1.7 s (initialization) $+$ 1.5 s (data processing) = 3.2 s. If the same task is performed by TensorFlow in the  32-bit FP arithmetic, the time is 1.8 s $+$ 0.2 s = 2.0 s. Hence, the time overhead coming with approximate operations is 37.5\%.

\subsection{Comparison with Other Similar Designs}

The results are compared with human-created ResNet networks of similar complexity as evolved CNNs. Parameters of ResNet-8 and ResNet-14 (utilizing the 8-bit exact multiplier) are depicted with crosses in Fig.~\ref{fig:final}. While ResNet-8 is dominated by several evolved CNNs, we were unable to evolve CNNs dominating ResNet-14. We further compared evolved CNNs with the CNNs optimized using the ALWANN method~\cite{Mrazek:iccad:2019}. ALWANN tries to identify the best possible assignment of an approximate multiplier to each layer of ResNet (i.e., different approximate multipliers can be assigned to different layers). For a good range of target energies, the proposed method produces better trade-offs than ALWANN.

Paper~\cite{Wistuba:19} reports 31 CNNs generated by various NAS methods and five CNNs designed by human experts. On CIFAR-10, the error is between 2.08\% and 8.69\%, the number of parameters is between 1.7 and 39.8 millions and the design time is between 0.3 and 22,400 GPU days. Our results are far from all these numbers as we address much smaller networks (operating with 8-bit multipliers) that must, in principle, show higher errors. However, paper~\cite{Tann2019} reports a number of human-created CNN hardware accelerators with the classification accuracy $80.77 - 81.53$\% on CIFAR-10, and with the total energy consumption $34.2-335.7\mu$J (energy of multiplications is not reported separately). These numbers are quite comparable with our results even under the conservative assumption that multiplication requires 20\% of the total energy of the accelerator.

\section{Conclusions}
\label{sec:conclusions}

We developed a multi-objective evolutionary design method capable of automated co-design of CNN topology and approximate multiplier(s). This is a challenging problem not addressed in the literature. On the standard CIFAR-10 classification benchmark, the CNNs co-optimized with approximate multipliers show excellent trade-offs between the classification accuracy and energy needed for multiplication in convolutional layers when compared with common ResNet CNNs utilizing 8-bit multipliers and CNNs optimized with the ALWANN method. Despite very limited computational resources, we demonstrated that it makes sense to co-optimize CNN architecture together with approximate arithmetic operations in a fully automated way. 

Our future work will be devoted to extending the proposed method, employing more computational resources, and showing its effectiveness on more complex problems instances. In particular, we will extend the CGP array size whose current setting was chosen to be comparable with the ALWANN method. It also seems that we should primarily focus on optimizing the convolutional layers and leave the structure of fully connected layers frozen for the evolution.

\subsubsection*{Acknowledgements}
%\vspace{3pt}
This work was supported by the Czech science foundation project 21-13001S.
The computational experiments were supported by The Ministry of Education, Youth and Sports from the Large Infrastructures for Research, Experimental Development and Innovations project ``e-Infrastructure CZ – LM2018140''.

\bibliographystyle{splncs04}
\bibliography{eurogp} 

\end{document}

%% file: nas_alg.tex
\begin{algorithmic}[1]
\State $P \gets$ initial\_population(); $g \gets 0$
\State training\_evaluation($P, E_{train}, D_{train}, D_{test}$)
%\State $g \gets 0$
\Repeat
\State $P' \gets$ replicate($P$); $O \gets$ mutate($P'$)
%\State $O \gets$ mutate($P'$)
\State training\_evaluation($O, E_{train}, D_{train}, D_{test}$)
\State $R \gets P \cup O$; $P \gets \emptyset$
%\State $P \gets \emptyset$
\While {$|P| \neq population\_size$}
\State $PF \gets non\_dominated(R)$
\If{$|P\cup PF| \leq population\_size$}
\State $P \gets P \cup PF$
\Else
\State $n \gets |PF \cup P| - population\_size$
\State $P \gets P \cup crowding\_reduce(PF, n)$
\EndIf
\State $R \gets R \setminus PF$
\EndWhile
\State $g \gets$ $g + 1$
\Until{stop\_criteria\_satisfied()}
\State training\_evaluation($P, E_{retrain}, D_{retrain}, D_{test}$)
\State \Return ($P$)
\end{algorithmic}

%% file: tab_all.tex
\begin{tabular}{cccrrllc} 
\toprule
\textbf{Method}                                & \multicolumn{2}{c}{\textbf{Accuracy}} & \multicolumn{1}{c}{\textbf{Energy}} & \multicolumn{1}{c}{\textbf{Mults}} &  & \textbf{Approx}.    & \multicolumn{1}{l}{\textbf{Energy of}}  \\
\multicolumn{1}{l}{}                           & Final & Estimated                     & {[}uJ]                              & \multicolumn{1}{c}{$\times 10^6$ } &  & \textbf{mult. ID}   & \textbf{1 mult. [pJ]}                   \\ 
\midrule
\multirow{14}{*}{Proposed} & 83.98 & 78.01 & 14.88 uJ & 30.9 &  & mul8u\_JD & 0.48 pJ \\
   & 83.50 & 76.88   & 13.82 uJ  & 30.9  &  & mul8u\_C1           & 0.45 pJ \\
   & 83.18 & 78.14   & 10.76 uJ  & 28.5  &  & mul8u\_GR           & 0.38 pJ \\
   & 83.01 & 77.02   & 6.79 uJ   & 22.9  &  & mul8u\_M1           & 0.30 pJ \\
   & 82.53 & 75.85   & 9.22 uJ   & 31.7  &  & mul8u\_85Q          & 0.29 pJ \\
   & 82.15 & 77.66   & 11.48 uJ  & 20.5  &  & mul8u\_JFF${}^{*}$  & 0.56 pJ \\
   & 81.03 & 75.54   & 2.23 uJ   & 7.7   &  & mul8u\_85Q          & 0.29 pJ \\
   & 79.55 & 62.32   & 0.14 uJ   & 27.2  &  & mul8u\_KX           & 0.01 pJ \\
   & 79.20 & 69.13   & 1.05 uJ   & 6.8   &  & mul8u\_2N4          & 0.15 pJ \\
   & 78.64 & 66.28   & 4.54 uJ   & 8.1   &  & mul8u\_JFF${}^{*}$  & 0.56 pJ \\
   & 77.66 & 59.12   & 0.14 uJ   & 26.8  &  & mul8u\_KX           & 0.01 pJ \\
   & 77.60 & 67.21   & 1.68 uJ   & 5.7   &  & mul8u\_M1           & 0.30 pJ \\
   & 76.73 & 60.88   & 0.03 uJ   & 4.9   &  & mul8u\_KX           & 0.01 pJ \\
   & 74.34 & 44.37   & 0.05 uJ   & 3.2   &  & mul8u\_8DU          & 0.02 pJ \\ 
\midrule
\multicolumn{1}{l}{\multirow{2}{*}{ALWANN}}    & 85.81 &  \multicolumn{1}{l}{ResNet-14}                     & 19.76 uJ                            & 35.3                               &  & mul8u\_JFF${}^{*}$  & 0.56 pJ                                 \\
\multicolumn{1}{l}{}                           & 82.85 &  \multicolumn{1}{l}{ResNet-8}                      & 11.84 uJ                            & 21.2                               &  & mul8u\_JFF${}^{*}$  & 0.56 pJ                                 \\
\bottomrule
\end{tabular}